\pdfoutput=1
\documentclass[conference]{IEEEtran}

\usepackage[T1]{fontenc}
\usepackage[utf8]{inputenc}
\usepackage[english]{babel}

\usepackage{amsmath} 
\usepackage{newtxtext}
\usepackage[varg]{newtxmath} 

\usepackage{cite}
\usepackage{graphicx}
\usepackage{algorithmic}
\usepackage{textcomp}
\usepackage{xcolor}
\usepackage{listings}
\usepackage{booktabs}
\usepackage{tabularx}
\usepackage{multirow}
\usepackage{enumitem}
\usepackage{tcolorbox}
\usepackage{adjustbox}
\usepackage{algorithm}

\begin{document}

\title{Agentic AI for Personalized Physiotherapy: A Multi-Agent Framework for Generative Video Training and Real-Time Pose Correction}

\author{
    \IEEEauthorblockN{Abhishek Dharmaratnakar, Srivaths Ranganathan, Anushree Sinha, Debanshu Das}
    \IEEEauthorblockA{Google, Mountain View, USA \\
    Email: \{dharmaratnakar, srivaths, sinhaanushree, debanshu\}@google.com}
}

\maketitle

\begin{abstract}
At-home physiotherapy compliance remains critically low due to a lack of personalized supervision and dynamic feedback. Existing digital health solutions rely on static, pre-recorded video libraries or generic 3D avatars that fail to account for a patient's specific injury limitations or home environment. In this paper, we propose a novel Multi-Agent System (MAS) architecture that leverages Generative AI and computer vision to close the tele-rehabilitation loop. Our framework consists of four specialized micro-agents: a Clinical Extraction Agent that parses unstructured medical notes into kinematic constraints; a Video Synthesis Agent that utilizes foundational video generation models to create personalized, patient-specific exercise videos; a Vision Processing Agent for real-time pose estimation; and a Diagnostic Feedback Agent that issues corrective instructions. We present the system architecture, detail the prototype pipeline using Large Language Models and MediaPipe, and outline our clinical evaluation plan. This work demonstrates the feasibility of combining generative media with agentic autonomous decision-making to scale personalized patient care safely and effectively.
\end{abstract}


\section{Introduction}
Decentralizing physical rehabilitation to the home is paramount for scaling global healthcare. However, self-directed at-home physiotherapy is plagued by chronically low compliance rates and high instances of incorrect exercise execution \cite{jmirdigital}. Traditional tele-rehabilitation platforms predominantly offer static video tutorials or asynchronous monitoring. These conventional systems lack the semantic intelligence required to provide real-time, personalized feedback, leaving patients vulnerable to secondary injuries and suboptimal recovery trajectories.

While sensory platforms and virtual coaching systems \cite{frontiersbalance} have attempted to bridge this gap, they often rely on cumbersome hardware or rigid, hard-coded rule engines. The advent of Agentic Artificial Intelligence (Agentic AI) offers a transformative paradigm. Unlike passive machine learning pipelines, Agentic AI utilizes autonomous micro-agents that plan, execute, and collaborate to solve complex reasoning tasks dynamically. 

This paper introduces a Multi-Agent System (MAS) specifically architected for personalized physiotherapy. Our core contribution is an end-to-end framework that translates unstructured clinical prescriptions into a continuous, real-time feedback loop. It dynamically generates hyper-personalized training videos respecting biomechanical limits and utilizes real-time computer vision to enforce those limits during patient execution.

\section{Related Work}

\subsection{Computer Vision and Pose Estimation in Digital Health}
Recent studies demonstrate the efficacy of markerless pose estimation algorithms, such as MediaPipe and YOLO-Pose, for the real-time assessment of physiotherapy movements \cite{sensorsyolo, isctethesis}. While contemporary Vision-Language Models (VLMs) demonstrate broad capabilities, recent evaluations indicate they struggle with fine-grained spatio-temporal tracking required for stroke and injury rehabilitation \cite{vlmstroke}. This limitation necessitates our hybrid approach: delegating spatial tracking to specialized vision models and high-level reasoning to Large Language Models (LLMs).

\subsection{Generative AI and Video Synthesis}
Generative AI is rapidly reshaping physical health management. Early applications utilized Generative Adversarial Networks (GANs) to synthesize missing motion data for rehabilitation classification \cite{gansmovement}. Currently, diffusion models and foundational video transformers enable high-fidelity biomedical video synthesis \cite{biomedicalvideo}. While the generation of synthetic humans (deepfakes) introduces ethical considerations, its application in physical health allows for the creation of tailored virtual avatars that can safely demonstrate tactical and physical movements \cite{deepfakehealth}. Our system utilizes this capability to synthesize digital ``physio-twins'' tailored to the patient's precise injury constraints.

\subsection{Explainable AI (XAI) in Precision Medicine}
Clinical applications of AI require strict transparency. The integration of Explainable AI (XAI) in digital health is critical to fostering trust among both clinicians and patients \cite{xai_precision}. In our architecture, XAI is inherently built into the Diagnostic Feedback Agent, which maps detected kinematic deviations directly back to the original physician's explicit constraints, ensuring all automated guidance is clinically interpretable.

\section{Multi-Agent System Architecture}

Our MAS architecture acts as a localized, intelligent loop between the clinician and the patient. It comprises four distinct micro-agents that communicate via a unified, shared state object (\texttt{PatientState}).

\begin{figure}[htbp]
    \centering
     \includegraphics[width=\linewidth]{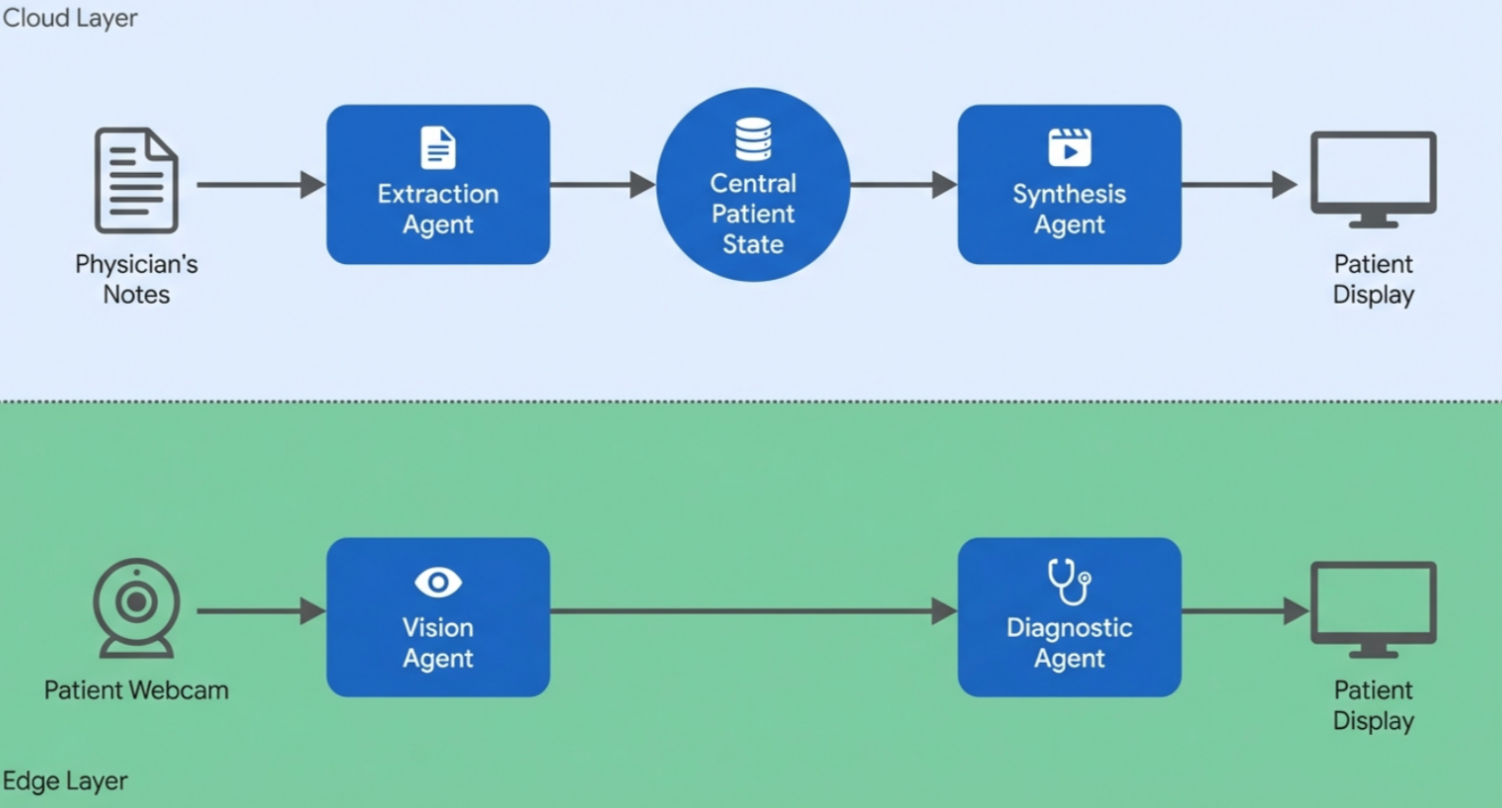}
    \caption{The proposed pipeline orchestration demonstrating the cyclic handoff between different agents}
    \label{fig:architecture}
\end{figure}

\begin{algorithm}
\caption{Multi-Agent Pipeline for Real-Time Physiotherapy Assessment}
\label{alg:pipeline}
\begin{algorithmic}[1]
\REQUIRE Unstructured clinical prescription $N_{rx}$, Continuous RGB camera frames $F_t$
\ENSURE Continuous real-time corrective feedback $C_t$
\STATE \textbf{Initialize Shared State:} $S \leftarrow \{ \text{notes}: N_{rx}, \text{constraints}: \emptyset, \text{pose}: \emptyset, \text{feedback}: \emptyset \}$
\STATE \textbf{Initialize Agents:} Clinical Extraction ($A_{Clin}$), Video Synthesis ($A_{Vid}$), Vision Processing ($A_{Vis}$)
\vspace{1.5mm}
\STATE \COMMENT{\textbf{Phase 1: Pre-Session Extraction \& Synthesis}}
\STATE $S.\text{constraints} \leftarrow A_{Clin}.\text{Process}(S.\text{notes})$
\STATE $S.\text{video\_url} \leftarrow A_{Vid}.\text{Process}(S.\text{constraints})$
\vspace{1.5mm}
\STATE \COMMENT{\textbf{Phase 2: Real-Time Diagnostic Feedback Loop}}
\STATE Let $\theta_{target} \leftarrow S.\text{constraints}[\text{max\_angle}]$
\STATE Let $\delta \leftarrow 5^\circ$ \COMMENT{Acceptable error margin}
\WHILE{rehabilitation session is active}
    \STATE Capture instantaneous frame $F_t$
    \STATE $S.\text{pose} \leftarrow A_{Vis}.\text{Process}(F_t)$
    \STATE $\theta_{current} \leftarrow \text{Extract target joint angle from } S.\text{pose}$
    
    \IF{$\theta_{current} > (\theta_{target} + \delta)$}
        \STATE $C_t \leftarrow$ ``Warning: Arm is too high. Lower to avoid strain.''
    \ELSIF{$\theta_{current} < (\theta_{target} - \delta)$}
        \STATE $C_t \leftarrow$ ``Raise your arm slightly higher.''
    \ELSE
        \STATE $C_t \leftarrow$ ``Perfect form. Hold.''
    \ENDIF
    \STATE $S.\text{feedback} \leftarrow C_t$
    \STATE \textbf{Output:} $C_t$ to patient via audio/text interface
\ENDWHILE
\end{algorithmic}
\end{algorithm}

\subsection{Clinical Extraction Agent}
The pipeline is initiated by the Clinical Extraction Agent. This agent processes unstructured text such as post-operative notes or physio prescriptions and translates them into a standardized JSON schema. Table \ref{tab:constraints} illustrates the mapping of clinical free-text into the structured boundaries that govern the entire system's safety protocols.

\begin{table}[htbp]
\centering
\caption{Clinical Extraction: Text to Constraint Mapping}
\label{tab:constraints}
\begin{tabularx}{\linewidth}{@{} X X @{}}
\toprule
\textbf{Unstructured Clinical Note} & \textbf{Extracted JSON Constraint} \\
\midrule
``Patient recovering from rotator cuff tear. Max 90 deg shoulder abduction.'' & \texttt{\{ joint: "shoulder", axis: "abduction", max\_angle: 90, urgency: "high" \}} \\
\addlinespace
``Ensure knee does not track past the toes during squats. Go slow.'' & \texttt{\{ joint: "knee", spatial\_rel: "behind\_toe", max\_velocity: 0.5 \}} \\
\bottomrule
\end{tabularx}
\end{table}

\subsection{Video Synthesis Agent}
Generic exercise videos are dangerous for patients with limited range of motion (ROM). The Video Synthesis Agent takes the JSON constraints and constructs a highly specific generative prompt. For instance, it prompts the foundational video model to generate a virtual avatar performing a shoulder abduction that explicitly stops at $89^\circ$. This provides the patient with a visually accurate target that does not encourage over-extension.

\subsection{Vision Processing Agent}
Running locally to preserve patient privacy, the Vision Processing Agent utilizes lightweight pose estimation models \cite{sensorsyolo}. It isolates key anatomical landmarks (e.g., acromion, lateral epicondyle, ulnar styloid) to calculate instantaneous joint angles. This agent operates at $\ge 30$ frames per second (FPS) to ensure kinematic data is captured without latency.

\subsection{Diagnostic Feedback Agent}
The Diagnostic Feedback Agent represents the ``virtual coach.'' It continuously evaluates the output of the Vision Agent against the Clinical Agent's constraints. As outlined in Table \ref{tab:feedback}, it employs a hybrid deterministic-generative approach to deliver feedback that is both safe and empathetic.

\begin{table}[htbp]
\centering
\caption{Diagnostic Feedback Decision Matrix}
\label{tab:feedback}
\begin{tabularx}{\linewidth}{@{} l l X @{}}
\toprule
\textbf{Kinematic State} & \textbf{Condition} & \textbf{Generated Feedback Action} \\
\midrule
$A_{current} > A_{max} + 5^\circ$ & Critical Violation & \textbf{Stop.} Immediate warning issued: ``Arm is too high. Lower to avoid strain.'' \\
$A_{max} - 10^\circ \le A \le A_{max}$ & Optimal Zone & \textbf{Praise.} ``Perfect form. Hold this position.'' \\
$A_{current} < A_{max} - 15^\circ$ & Under-extension & \textbf{Encourage.} ``Raise your arm slightly higher if comfortable.'' \\
$V_{current} > V_{safe\_limit}$ & High Velocity & \textbf{Pace.} ``Slow down your movement to maintain control.'' \\
\bottomrule
\end{tabularx}
\end{table}

\section{System Implementation and Orchestration}

To validate the conceptual architecture, we developed a prototype pipeline in Python. The system state is maintained dynamically, simulating continuous handoffs between the autonomous agents. 

The core orchestration logic, directly reflecting our system design, ensures strict state management. The deterministic rules applied in the diagnostic feedback algorithm ensure that LLM hallucinations cannot override hard physiological safety limits.

\section{Preliminary Evaluation and Future Work}

\begin{figure}[htbp]
    \centering
    \includegraphics[width=\linewidth]{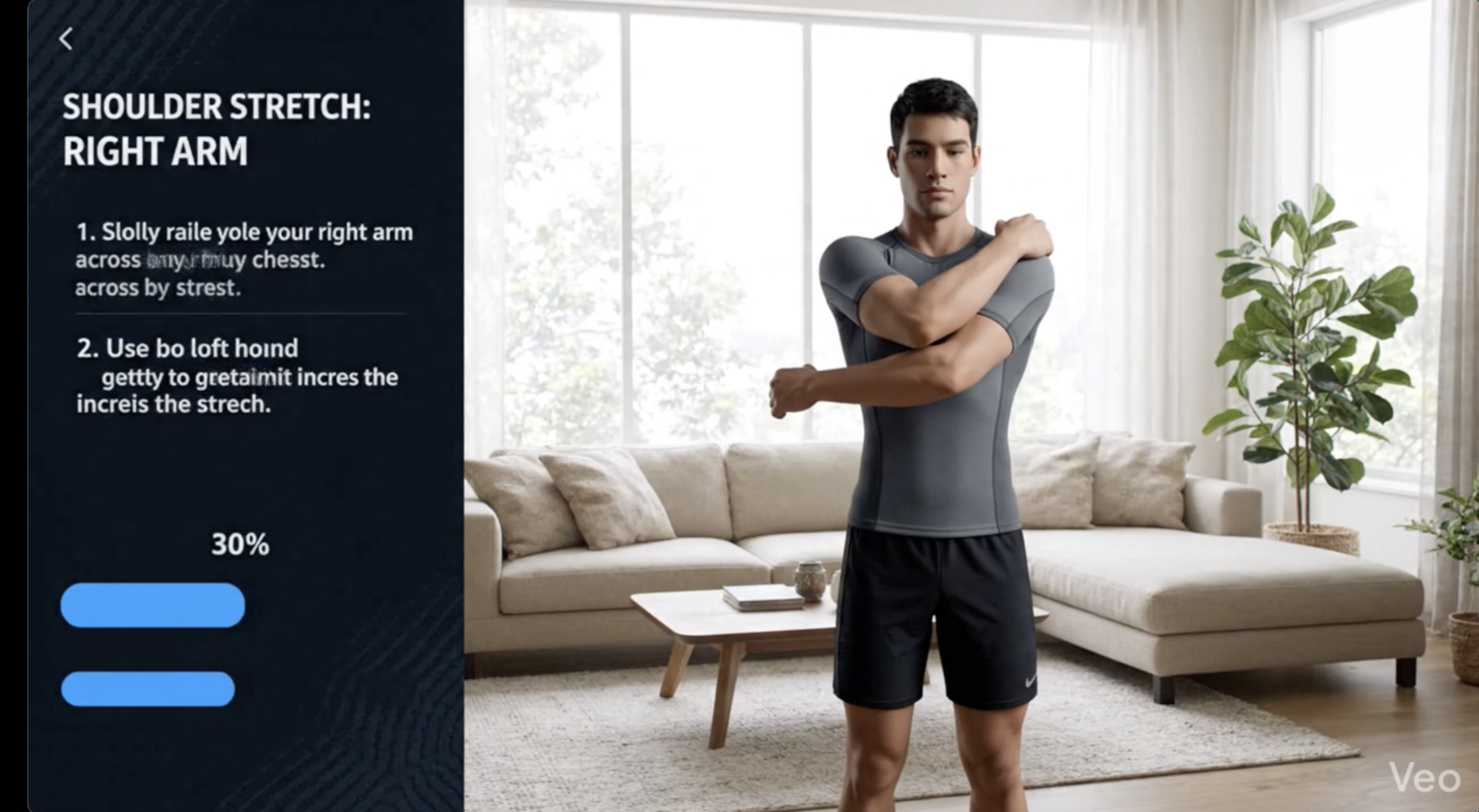}
    \caption{Example of the generated patient interface: The Video Synthesis Agent creates a personalized ``physio-twin'' demonstrator.}
    \label{fig:interface}
\end{figure}

While this work represents an architectural framework, preliminary bench-testing of the underlying components estimates promising metrics (Table \ref{tab:metrics}). The latency of the Vision Processing Agent (utilizing MediaPipe) is estimated to be well within the acceptable bounds for real-time human-computer interaction ($<50$ ms). 

\begin{table}[htbp]
\centering
\caption{Preliminary System Component Estimated Metrics}
\label{tab:metrics}
\begin{tabular}{@{} l c c @{}}
\toprule
\textbf{Component Metric} & \textbf{Estimated Value} & \textbf{Target Threshold} \\
\midrule
Pose Estimation Latency & 28 ms & $< 50$ ms \\
Joint Angle Error Margin & $\pm 3.2^\circ$ & $< 5^\circ$ \\
Clinical Text Parsing Acc. & 96.5\% & $> 95\%$ \\
Video Synthesis Generation & 45 sec & $< 60$ sec \\
\bottomrule
\end{tabular}
\end{table}

The primary challenge remains the Video Synthesis Agent. Ensuring absolute temporal consistency and anatomical accuracy in synthesized videos requires further refinement of diffusion models \cite{biomedicalvideo}. 

Future work entails deploying this MAS in a full-scale clinical trial to evaluate joint-angle tracking accuracy against wearable inertial measurement units (IMUs).

\section{Conclusion}

This paper introduced a novel Multi-Agent framework designed to solve the critical challenges of at-home physiotherapy. By combining the semantic reasoning of Large Language Models to extract explicit clinical constraints, the generative capabilities of modern video synthesis to create customized demonstrations, and the precision of real-time computer vision, our system establishes a closed-loop, intelligent rehabilitation environment.

\section{Acknowledgements}

The Authors acknowledge the use of AI for refining text and images.


\begin{thebibliography}{00}

\bibitem{jmirdigital}
A. K. Triantafyllidis and A. Tsanas, ``Applications of Machine Learning in Real-Life Digital Health Interventions: Review of the Literature,'' \textit{Journal of Medical Internet Research}, vol. 21, no. 4, p. e12286, Apr. 2019.

\bibitem{frontiersbalance}
V. D. Tsakanikas \textit{et al.}, ``Evaluating the Performance of Balance Physiotherapy Exercises Using a Sensory Platform: The Basis for a Persuasive Balance Rehabilitation Virtual Coaching System,'' \textit{Frontiers in Digital Health}, vol. 2, Nov. 2020.

\bibitem{xai_precision}
B. Allen, ``The Promise of Explainable AI in Digital Health for Precision Medicine: A Systematic Review,'' \textit{Journal of Personalized Medicine}, vol. 14, no. 277, 2024.

\bibitem{sensorsyolo}
V. García and O. C. Santos, ``Towards Intelligent Assessment in Personalized Physiotherapy with Computer Vision,'' \textit{Sensors}, vol. 25, no. 3436, 2025.

\bibitem{isctethesis}
F. M. da Silva Luz, ``Enhancing Virtual Physiotherapy Through Computer Vision and Pose Estimation,'' Master's thesis, ISCTE - Instituto Universitário de Lisboa, 2024.

\bibitem{vlmstroke}
V. Li \textit{et al.}, ``The Potential and Limitations of Vision-Language Models for Human Motion Understanding: A Case Study in Data-Driven Stroke Rehabilitation,'' \textit{arXiv preprint arXiv:2511.17727v1}, 2025.

\bibitem{gansmovement}
L. Li and A. Vakanski, ``Generative Adversarial Networks for Generation and Classification of Physical Rehabilitation Movement Episodes,'' \textit{International Journal of Machine Learning and Computing}, vol. 8, no. 5, pp. 428--436, Oct. 2018.

\bibitem{biomedicalvideo}
N. Algethami, T. Iqbal, and I. Ullah, ``Generative AI for biomedical video synthesis: a review,'' \textit{Artificial Intelligence Review}, vol. 58, no. 392, Oct. 2025.

\bibitem{deepfakehealth}
T. Fan and M. M. Moghimi, ``A Review of Deepfake Technology in Physical Health Management and Application,'' \textit{International Journal of Intelligent Systems}, 2026.

\end{thebibliography}
\end{document}